\newif\ifarxiv \arxivtrue

\ifarxiv
  \PassOptionsToPackage{capitalize}{cleveref}
  \PassOptionsToPackage{section}{placeins}
  \documentclass{applemlr}
\usepackage{silence}
\makeatletter
\patchcmd{\wrong@fontshape}{\@gobbletwo}{}{}{}
\makeatother
\makeatletter
\AtBeginDocument{
  \urlstyle{sf}
  
}
\makeatother

\else
  \documentclass{article} 
  \usepackage{iclr2027_conference,times}
\fi

\usepackage{amsmath,amsfonts,bm}

\def\eqref#1{equation~\ref{#1}}

\def\1{\bm{1}}

\DeclareMathAlphabet{\mathsfit}{\encodingdefault}{\sfdefault}{m}{sl}
\SetMathAlphabet{\mathsfit}{bold}{\encodingdefault}{\sfdefault}{bx}{n}

\usepackage{titletoc}
\ifarxiv\else
  \usepackage{hyperref}
  \definecolor{darkorange}{HTML}{BF572E}
  \hypersetup{
    colorlinks=true,
    linkcolor=darkorange,
    citecolor=darkorange,
    urlcolor=darkorange
  }
  \usepackage{microtype}
\fi
\usepackage{url}

\usepackage{graphicx}
\usepackage{wrapfig}
\usepackage{tikz}
\usetikzlibrary{arrows.meta,shapes.geometric}

\usepackage{multirow}
\usepackage{booktabs}
\usepackage{graphicx}
\usepackage{subcaption}
\usepackage{tikz}
\usetikzlibrary{shapes.geometric, positioning, fit}
\usepackage{algorithm}
\usepackage{algpseudocode}
\usepackage[most]{tcolorbox}
\usepackage{longtable}

\DeclareMathOperator{\reshape}{reshape}
\DeclareMathOperator{\mean}{mean}

\definecolor{linered}{HTML}{E41A1C}
\definecolor{fillred}{HTML}{FCE8E8}
\definecolor{fillgray}{HTML}{EEEEEE}
\definecolor{linegray}{HTML}{444444}

\newtcolorbox{graytitlebox}[1]{%
  enhanced,                
  rounded corners,         
  arc=1.5mm,                 
  colback=gray!15,         
  colframe=gray!50!black,  
  boxrule=0.5pt,           
  title={#1},              
  fonttitle=\bfseries,     
  coltitle=white,          
  breakable,               
  boxsep=2mm,              
}

\definecolor{cCost}{HTML}{4A2A82}
\definecolor{cPrec}{HTML}{3A8C52}
\definecolor{cRank}{HTML}{E6A845}
\definecolor{cDepth}{HTML}{595959}
\definecolor{gGQAbf}{HTML}{7EC19B}
\definecolor{gGQAq}{HTML}{3A8C52}
\definecolor{gMLAbf}{HTML}{FED286}
\definecolor{gMLAq}{HTML}{E6A845}
\definecolor{gKIVI}{HTML}{5497C1}
\tikzset{gsolid/.style={line width=0.9pt}, gdash/.style={line width=0.9pt, dash pattern=on 1.8pt off 1.1pt}}

\usepackage{booktabs}
\usepackage{multirow}
\usepackage{amsmath}
\usepackage{amssymb}
\ifarxiv\else
  \usepackage[capitalize,noabbrev]{cleveref}
  \usepackage[section]{placeins}
\fi

\crefname{appsec}{Appendix}{Appendices}
\Crefname{appsec}{Appendix}{Appendices}

\title{Lossy Compressive Text Autoencoders}

\ifarxiv
  \metadata[VS]{\sffamily Work done while interning at Apple.}
  \metadata[AK]{\sffamily Work done while at Apple.}
  \author{Vinko Sabolčec}
\author{Angelos Katharopoulos}
\author{David Grangier}

\affiliation{Apple}
  \abstract{Our work explores learning a compressed latent representation of text, at the intersection of data compression and representation learning. We propose an autoencoder architecture that performs residual downscaling and upscaling of hidden representations along the time axis,
with a residual low-dimension discrete bottleneck. We analyze our approach for different quantization methods, training objectives, and datasets. For different levels of compression,
we evaluate the similarity between the original and reconstructed text
both at the surface-level (BLEU) and at the semantic-level (LLM-based judge). Additionally, we evaluate our models on downstream question-answering and semantic text similarity benchmarks. Our approach results in compressed representations which are on par with lossless text compression algorithms at 2.24 bits per byte on web text data, while having good reconstruction and downstream task performance.
}
  \metadata[Correspondence]{\sffamily
  Vinko Sabolčec: \url{vinko.sabolcec@epfl.ch};
  David Grangier: \url{grangier@apple.com}
  }
\else
  \author{Vinko Sabolčec, Angelos Katharopoulos, David Grangier\\ Apple MLR}
\fi

\begin{document}

\maketitle

\section{Introduction}

The ability to represent information efficiently is at the heart of both data compression and artificial intelligence. Whether the goal is compression per se or generating the next-token prediction using a model trained on a large corpus, the underlying challenge is the same: learning to represent data in a way that preserves its essential meaning in the smallest possible space~\citep{lmiscompression}. While the current language models (LMs) have demonstrated the ability to learn highly useful representations, these representations are often massive in scale compared to the raw information they represent. Furthermore, compression is usually treated as a separate utility which remains disconnected from the structure of the representation space.

In this work, we explore the joint goal of representation learning and data compression using neural models. While there is existing work focusing on compression, such as lossless text compressors using entropy coding on the next-token probability distribution~\citep{ts_zip,nacrith} and context compression for faster inference~\citep{bigbird,e2econtextcompression}, or representation learning, such as embedding~\citep{qwen3embedding} or sequence-to-sequence models~\citep{sonar}, these methods treat representation learning and data compression as disconnected tasks. On the other hand, we explore a paradigm of learning expressive representations that are compressed by design. In other words, we ask: how can we train a general-purpose lossy text reconstruction model that preserves semantic information while achieving a high compression rate?

In summary, we make the following contributions:
\begin{itemize}
    \item We propose a Transformer-based~\citep{transformer} text autoencoder (AE) architecture that performs residual downscaling and upscaling of hidden representations along the time axis, uses a residual bottleneck projection, and quantization to produce compressed latent representations. We achieve 2.24 bits per byte compression rate using a 4.3B parameter autoencoder model on web text data which, to the best of our knowledge, is the highest compression rate for a text autoencoder model in literature.
    \item We explore different quantization methods, training objectives, and datasets for autoencoder models with compressed latent representations and analyze their effects on the trained models, as well as evaluate the text reconstructions and representations on downstream question-answering and semantic text similarity tasks. We find that there is a reconstruction-representation trade-off, with our text autoencoder models favoring reconstruction over representation.
    \item We show that common text similarity metrics, such as BLEU~\citep{bleu}, Rouge~\citep{rouge}, chrF~\citep{chrf}, and BERTScore~\citep{bertscore}, do not reliably measure semantic similarity between two texts, and we introduce and evaluate an LLM judge based on an adaptation of Multidimensional Quality Metrics (MQM;~\citealt{mqm,mqm2}).
\end{itemize}

\section{Related Work}

\textbf{Lossless Text Compression.} Common traditional approaches for lossless text compression include dictionary-based methods, such as the Lempel-Ziv algorithms and their variants~\citep{lz77,lz78}, entropy coding methods~\citep{huffman-code,arithmetic-code,ans}, and transforms~\citep{rle,mtf,burrows-wheeler-transform}, often used in combination to reduce entropy and make the text more compressible~\citep{deflate,zstandard}. On the other hand, neural-based lossless text compression methods use neural networks to model sequences and compress the data using entropy coding~\citep{cmix,nncpv1,nncpv2,llmzip}, with research mainly focusing on improving runtime efficiency and compression rates~\citep{ts_zip,finezip,nacrith}. In the context of text compression, we explore lossy compression using neural-based autoencoder models which encode and reconstruct text that is semantically equivalent to the input text, and their rate-distortion trade-off.

\textbf{Context Compression.} Improving efficiency of Transformer models over long text sequences was explored through the change of attention patterns to retain the expressivity while reducing the quadratic complexity~\citep{longformer,etc,bigbird}, and alternative architectures whose state and computation does not grow with context length~\citep{lstm,linearattention,ssm,mamba,gdn}. Recent context compression research also explored representation of sequences using fewer latent tokens, such as through OCR~\citep{deepseek-ocr} or text autoencoding-based methods~\citep{ccc,e2econtextcompression}, achieving up to 40x compression of the token sequence. Similar to autoencoding-based context compression methods, we use a Transformer-based autoencoder architecture to compress the model inputs along the time dimension into a compact latent representation from which the original output text can be reconstructed. We propose a new architecture which results in larger context compression rates.

\textbf{Text Embedding Models.} Text embedding models encode a token sequence into a meaningful latent representation, commonly used for tasks such as semantic similarity, document retrieval, clustering, and classification~\citep{mteb}. Embedding models are usually obtained from Transformer-based language models trained on the unmasking~\citep{bert} or the next-token prediction task~\citep{qwen3embedding}, with the latent representation obtained by mean-pooling the latent token representations, or by adding a dedicated token to the sequence, such as the beginning-of-sequence or the end-of-sequence token. To further improve performance on downstream tasks, models are finetuned on task-specific data. While the goal of our work is not to create a text embedding model, we compare the latent representations of our compressive autoencoders to existing embedding models, as well as their text reconstruction capabilities.

\textbf{Text Reconstruction from Latent Representations.} \citet{vec2text-cornell} showed that latent representations from text embedding models which were not originally trained for text reconstruction can be used to obtain text that is semantically close to the original through iterative text correction and generation. On the other hand, (Omni-)SONAR~\citep{sonar,omnisonar} and \citet{vec2text-google} explore text reconstruction using autoencoder models where both the encoder and the decoder are trained on the text reconstruction task. Similar to (Omni-)SONAR and \citet{vec2text-google}, we train a text autoencoder, while exploring latent representation compression and additional supervision objectives.

\section{Method}

\subsection{Experimental Setup}

In this section, we train 0.6B parameter Transformer-based autoencoder models with hidden dimension 768 and gated MLPs in the \texttt{bfloat16} data type (16 bits per scalar). We use the cross-entropy loss, with model input and target tokens sampled from the Nemotron High-quality subset~\citep{nemotron} and tokenized using the Qwen3-0.6B tokenizer~\citep{qwen3}. We use a batch size of 256K tokens, a peak learning rate of $2\cdot10^{-4}$ with a 5\% linear warmup followed by cosine decay, and the Adam optimizer~\citep{adam}. Following our initial experiments on a 1.4B parameter model, we allocate 24 tokens per parameter, at which we start to observe the loss converge, although we note that longer training continues to improve the loss. For evaluation of semantic similarity between the input and the reconstructed text, we report the 95\% confidence intervals of the BLEU score (ranging between 0 and 1;~\citealt{bleu}) and the Qwen3.6-35B-A3B LLM judge score (ranging between 1 and 10;~\citealt{qwen36_35b_a3b}) with thinking enabled on 1000 validation documents following the training data distribution. Borrowing the terminology from the machine translation field, we use the input text as the reference and sample the reconstructions from our trained models using a temperature of 0.6 as the hypothesis, and compute the final score as the mean of the per-sample metric scores. We report the \textbf{best result in bold} and the \underline{second best result underlined}. Beyond this, we provide more evaluation details in Section~\ref{sec:evaluation}.

\subsection{Architecture}\label{sec:arch}

\textbf{Base Architecture.} As our baseline, we use a Transformer-based autoencoder architecture which consists of: (1) Transformer encoder layers with the bidirectional attention mechanism, (2) pooling and bottleneck projection operations where the encoder output sequence is transformed into the final latent representation by either Mean-pooling or CLS-token pooling, followed by dimensionality-reducing MLP projection, and (3) Transformer decoder layers where an autoregressive (AR) Transformer reconstructs the text by cross-attending to the pooled and projected latent representation. Variants of this architecture are found in prior work, such as~\citet{ccc} and~\citet{e2econtextcompression}.

\textbf{Multi-resolution Autoencoder Architecture.} Inspired by~\citet{unet}, we propose the multi-resolution autoencoder architecture which introduces the following changes: (1) progressive scaling where the intermediates are progressively downscaled by the encoder and upscaled by the upscaler along the time dimension, removing the pooling operation, (2) residual bottleneck projection which reduces the dimensionality while preserving the residual path, and (3) upscaler supervision, with a non-autoregressive (non-AR) loss from the next-token prediction head applied to the upscaled intermediates.
We show the multi-resolution autoencoder architecture in Figure~\ref{fig:architecture}.

We implement progressive scaling along the time axis by replacing the standard residual MLP operation in the Transformer layer with a residual sequence rescaling operation of factor~$F$ that uses MLP$_F$ as shown in Algorithm~\ref{alg:residual_rescale}, where MLP$_F$ is an MLP with input $X_{\text{in}} \in \mathbb{R}^{B \times L \times FD}$ and output $X_{\text{out}} \in \mathbb{R}^{B \times L \times D}$. For example, multi-resolution layer with MLP$_1$ is equivalent to the standard Transformer layer, while using MLP$_2$ reduces the sequence length by 2. Since the scaling factor is multiplicative, stacking $K$ multi-resolution layers of factors $F_1, \dots, F_K$ results in the total scaling factor of $F_{\text{total}}=\prod_{i=1}^K F_i$, which transforms the input input $X_{\text{in}} \in \mathbb{R}^{B\times F_{\text{total}}L\times D}$ to the output $X_{\text{out}} \in \mathbb{R}^{B \times L \times D}$. We implement the residual bottleneck projection using an algorithm equivalent to Algorithm~\ref{alg:residual_rescale}, with the only change being the axis along which the operation is performed.

\begin{figure*}[htb!]
    \centering
    \begin{minipage}{0.48\textwidth}
        \centering
        \resizebox{0.98\linewidth}{!}{
        \begin{tikzpicture}[
    font=\sffamily,
    thick,
    line width=1.5pt,
    draw=linered,
    token/.style={
        draw=linegray, 
        fill=fillgray, 
        rounded corners=5pt, 
        minimum width=1.3cm, 
        minimum height=1.3cm
    },
    label/.style={
        align=center, 
        font=\sffamily\Large, 
        text=black
    },
    blocktext/.style={
        align=center, 
        font=\sffamily\fontsize{13pt}{14pt}\selectfont, 
        text=black
    }
]

\foreach \i/\y in {1/2.8, 2/1.4, 3/0, 4/-1.4, 5/-2.8} {
    \node[token] (in\i) at (0,\y) {};
}
\node[label] at (0, 4.2) {Input\\[-2pt]Tokens};

\draw[fill=fillred, rounded corners=5pt] 
    (0.9, 3.4) -- (3.0, 0.9) -- (3.0, -0.9) -- (0.9, -3.4) -- cycle;
\node[blocktext] at (2.0, 0) {Encoder};

\node[token, fill=fillgray] (latent) at (3.9, 0) {};
\node[label] at (3.9, 2.7) {Latent\\[-2pt]Representation};

\draw[fill=fillred, rounded corners=5pt] 
    (4.8, 0.9) -- (6.7, 3.4) -- (6.7, -3.4) -- (4.8, -0.9) -- cycle;
\node[blocktext] at (5.75, 0) {Non-AR\\Upscaler};

\foreach \i/\y in {1/2.8, 2/1.4, 3/0, 4/-1.4, 5/-2.8} {
    \node[token] (up\i) at (7.6,\y) {};
}
\node[label] at (7.6, 4.2) {Upscaled\\[-2pt]Intermediates};

\draw[fill=fillred, rounded corners=5pt] 
    (8.5, 3.4) rectangle (10.6, -3.4);
\node[blocktext] at (9.6, 0) {AR\\Decoder};

\foreach \i/\y in {1/2.8, 2/1.4, 3/0, 4/-1.4, 5/-2.8} {
    \node[token] (out\i) at (11.5,\y) {};
}
\node[label] at (11.5, 4.2) {Reconstructed\\[-2pt]Tokens};

\end{tikzpicture}
        }
        \caption{Overview of the multi-resolution autoencoder architecture.}
        \label{fig:architecture}
    \end{minipage}%
    ~ \hfill
    \begin{minipage}{0.48\textwidth}
        \vspace{-24pt}
        \begin{algorithm}[H]
        \small
        \caption{Residual sequence rescale of factor $F$}\label{alg:residual_rescale}
            \begin{algorithmic}
                \State \hspace{-1.2em} \textbf{Input:} $X_{\text{in}} \in \mathbb{R}^{B \times FL \times D}$
                \State \hspace{-1.2em} \textbf{Output:} $X_{\text{out}} \in \mathbb{R}^{B \times L \times D}$
                \State \hspace{-1.2em} \textbf{Uses:} $\text{MLP}_{F}: X_{\text{in}} \in \mathbb{R}^{B \times L\times FD'} \to \mathbb{R}^{B \times L\times D'}$
                \vspace{4pt}
                \State $X' \gets \reshape \bigr(X_{\text{in}}, (B, \min(FL, L), -1, D)\bigl)$
                \State $X' \gets \mean \bigr(X', \text{dim}=-2, \text{keepdim}=\text{True}\bigl)$
                \State $Y \gets \text{MLP}_F\bigr(X_{\text{in}}\bigl)$
                \State $Y \gets \reshape\bigr(Y, (B, \min(FL, L), -1, D)\bigl)$
                \State $X_{\text{out}} \gets \reshape \bigr(X' + Y, (B, -1, D)\bigl)$
            \end{algorithmic}
        \end{algorithm}
    \end{minipage}%
\end{figure*}

\textbf{Experimental Results.} We compare the baseline architectures (Mean and CLS Pool AE) against our multi-resolution AE architecture at 128 sequence length and using the 384-dimensional bottleneck. We choose this  dimension such that it lies neither at the upper nor at the lower extrema of the evaluation metrics, see Figure~\ref{fig:arch_cont_dim} (Appendix~\ref{appendix:additional_experiments}). For our multi-resolution AE models, we use 14 encoder layers alternating $1.0$ and $2.0$ scaling factors 7 times (total scaling factor $128=2^7$), 14 upscaler layers alternating $0.5$ and $1.0$ scaling factors 7 times, and 14 decoder layers. For the experiment with the upscaler loss, the final loss is computed as the average between the upscaler and the decoder loss. Additionally, we include results for different upscaler and decoder loss weights in Table~\ref{table:upscaler_decoder_loss_comparison} (Appendix~\ref{appendix:additional_experiments}). For the Mean and CLS Pool baselines, we use an equivalent configuration consisting of 14 encoder layers and 28 decoder layers, and also include a comparison with our proposed residual bottleneck projection. In Table~\ref{table:arch_comparison}, we observe that Mean and CLS Pool baselines struggle to compress the sequence of 128 tokens, which is in accordance with prior work that does not go beyond compressing sequences of more than 40 tokens using one latent representation~\citep{ccc,e2econtextcompression}. Residual projection alone improves the Mean Pool and CLS Pool baselines, indicating the difficulty of learning the bottleneck MLP projection and near-identity being a good initialization. Lastly, our multi-resolution AE architecture with the added upscaler loss performs the best and second-best on BLEU and judge score, respectively, highlighting the effectiveness of our proposed changes in a context compression setting. We show the scaling of the architecture at different sequence lengths in Figure~\ref{fig:arch_seq_len} (Appendix~\ref{appendix:additional_experiments}).

\begin{table}[htb!]
\caption{BLEU and LLM judge score comparison of 0.6B parameter baseline and multi-resolution AE models trained on 128 sequence length, scaling factor of 128, and latent dimension 384.}
\label{table:arch_comparison}
\begin{center}
\small
\begin{tabular}{lrr}
\toprule
\multicolumn{1}{c}{\bf Architecture}   &\multicolumn{1}{c}{\bf BLEU $\uparrow$}   &\multicolumn{1}{c}{\bf Judge Score $\uparrow$}
\\
\midrule
Mean Pool AE & $0.12 \pm 0.01$ & $1.49 \pm 0.11$ \\
\hspace{1em} with Residual Projection & $0.77 \pm 0.01$ & $5.62 \pm 0.21$ \\
CLS Pool AE & $0.20 \pm 0.01$ & $1.83 \pm 0.13$ \\
\hspace{1em} with Residual Projection & \underline{$0.80 \pm 0.01$} & $\mathbf{5.90 \pm 0.21}$ \\
Multi-resolution AE & $0.77 \pm 0.01$ & $5.59 \pm 0.21$ \\
\hspace{1em} with Upscaler Loss & $\mathbf{0.91 \pm 0.01}$ & $\underline{5.78 \pm 0.20}$ \\

\bottomrule
\end{tabular}
\end{center}
\end{table}

\subsection{Latent Representation Quantization}\label{sec:quant}

\textbf{Quantization Approaches.} To further compress the latent representation, we introduce the quantization operation as the final step of the encoder, a standard practice in neural image~\citep{image_compression_1,image_compression_2} and audio compression~\citep{soundstream,encodec}. We evaluate both scalar and vector quantization approaches. Scalar quantization approaches map each scalar independently to its quantized value. For our experiments, we use finite scalar quantization (FSQ;~\citealt{fsq}). On the other hand, vector quantization approaches map the whole vector to its quantized value using a codebook. For our experiments, we explore residual vector quantization (RVQ;~\citealt{rvq}) which sequentially applies the vector quantization operation on the residual of the previous step, product residual vector quantization (PRVQ) which performs RVQ on latent representation subspaces by treating a single vector as separate vector chunks, and chunked residual vector quantization (CRVQ) where we use an encoder with the same number of layers but a smaller total scaling factor and a narrower residual bottleneck projection, with RVQ applied to each of these latent representations. Specifically, CRVQ chunking results in one narrower latent representation per chunk of tokens defined by the total scaling factor.

\textbf{Experimental Results.} We train 0.6B parameter multi-resolution AE models at sequence length 128 and 384 latent bottleneck dimension. We set the hyperparameters of each of the quantization methods to produce representations at approximately 9 bits per token to lie between the upper and lower extrema of the evaluation metrics. Specifically, we use FSQ with 8 values per scalar, RVQ with codebook size 4096 and 96 quantizers, PRVQ with codebook size 2048, 7 quantizers and 16 subspaces, and CRVQ with codebook size 2048, 7 quantizers, scaling factor 8 and latent dimension 24. For vector quantization approaches, we update the codebook as the exponential moving average of the quantizer inputs, use the commitment loss, and perform dead code rebalancing. Lastly, we include an experiment where we decrease the latent bottleneck dimension without using a quantizer as a baseline (Narrow Bottleneck). In Table~\ref{table:quant}, we observe that naively reducing the latent bottleneck dimension results in poor text reconstruction performance. Similarly, we observe that RVQ struggles to quantize high-dimensional vectors, which is improved by PRVQ. We see an improvement using FSQ and CRVQ, with CRVQ having the highest performance. We show additional CRVQ chunking hyperparameters in Table~\ref{table:crvq_chunks} (Appendix~\ref{appendix:additional_experiments}). These results highlight the importance of the structure of the latent representations that the quantizers can exploit to turn into discrete codes. For example, we can draw a parallel to gzip+kNN text classification~\citep{gzipknn} where the bzip2 compression algorithm shows larger compression rates than gzip but performs worse on a downstream classification task because bzip2 uses the Burrows-Wheeler transform~\citep{burrows-wheeler-transform}, permuting character order and discarding the structure that the distance metric needs.

\begin{table}[htb!]
\caption{BLEU and LLM judge score comparison of 0.6B parameter multi-resolution AE models trained at approximately 9 bits per token on 128 sequence length, scaling factor of 128, and latent dimension 384, except CRVQ which uses scaling factor 8 and latent dimension 24.}
\label{table:quant}
\begin{center}
\small
\begin{tabular}{llrrrr}
\toprule
\multicolumn{1}{c}{\bf Approach}    &\multicolumn{1}{c}{\bf Representation}     &\multicolumn{1}{c}{\bf Bits per Token $\downarrow$}   &\multicolumn{1}{c}{\bf BLEU $\uparrow$}   &\multicolumn{1}{c}{\bf Judge Score $\uparrow$}
\\ \midrule
No Quantizer & Continuous & $48.00$  & $0.91 \pm 0.01$ & $5.78 \pm 0.20$ \\
Narrow Bottleneck & Continuous & $12.00$ & $0.10 \pm 0.01$ & $1.58 \pm 0.11$ \\
\midrule
FSQ & Discrete & $9.00$ & $\underline{0.44 \pm 0.01}$ &  $\underline{2.41 \pm 0.13}$ \\
RVQ & Discrete & $9.00$ & $0.07 \pm 0.01$ & $1.44 \pm 0.10$ \\
PRVQ & Discrete & $9.63$ & $0.33 \pm 0.01$ & $2.13 \pm 0.11$ \\
CRVQ & Discrete & $9.63$ & $\mathbf{0.76 \pm 0.01}$ &  $\mathbf{3.25 \pm 0.14}$ & \\
\bottomrule
\end{tabular}
\end{center}
\end{table}

\begin{figure}[htb!]
\centering
\includegraphics[width=0.6\linewidth]{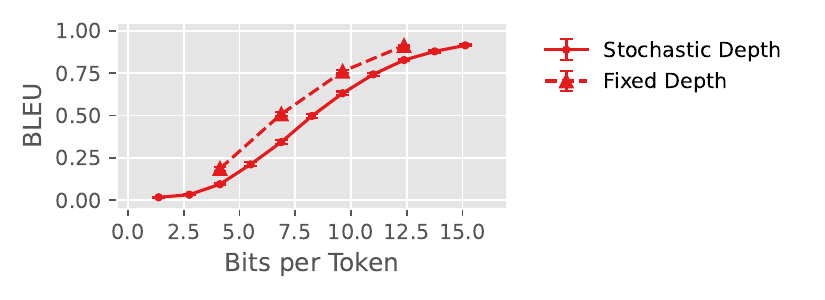}
\caption{BLEU score comparison of multi-resolution AE model trained with stochastic CRVQ depth, uniformly sampled between 1 and 11 during training, at different inference-time depths, with the fixed-depth baselines (depths 3, 5, 7, and 9).}
\label{fig:quant_vbr}
\vspace{-0.7em}
\end{figure}

\textbf{Rate-Distortion Trade-off.} To analyze the rate-distortion trade-off, in addition to training multiple models at fixed CRVQ depths, we explore stochastic depth training to explore models which can trade off between reconstruction and compression during inference. For stochastic CRVQ depth, we uniformly sample the number of CRVQ quantizers for each sequence in a batch during training. In Figure~\ref{fig:quant_vbr}, we observe that training using stochastic depth consistently degrades performance compared to fixed-depth training. As we approach the 17.2 bits per token needed to store the \mbox{Qwen3-0.6B} token IDs without any compression, BLEU score for both increases toward its maximum, indicating that the models primarily rely on copying the tokens inside the latent representations. This poses a significant problem for this class of models, as making the bottleneck smaller might help learn better representations, however, at the cost of poor text reconstruction.

\subsection{Training Supervision}\label{sec:supervision}

We explore whether the models can be improved through different types of supervision at the loss, masking, and dataset levels.

\textbf{Loss Functions.} In addition to the cross-entropy (CE) loss, we experiment with three loss functions: (1) bounded cross-entropy loss, BoundedCE${_{b}}$, where tokens with a loss smaller than $b$ receive no gradient so that model capacity is spent on high loss tokens, i.e. mistakes, (2) upweighting the loss of the a desired token class to $w$, WeightedClass$_{w}$, where we upweight the loss on numbers and proper nouns to invest more capacity on immutable information, and (3) including the language modeling loss, CE$_{LM}$, to improve predictions of the decoder, with the cross-attention not being used and instead a learned vector being added.

\textbf{Token Masking.} We mask the: (1) encoder inputs with the goal of learning better representations, and (2) decoder input prefixes as an alternative to the upscaler loss, both of which are studied in literature~\citep{bert,tsdae,retromae}. For encoder masking, we gradually cool down the proportion of the masked tokens at the end of training to adapt the model for inference. For decoder masking, we uniformly mask a prefix of the desired proportion of the sequences in the batch and keep the rest unmasked, without using the upscaler loss.

\textbf{Dataset Variants.} In addition to our dataset with matching inputs and targets (Copy), we introduce two datasets: (1) round-trip translations (RTT) generated by machine translating data to German and back to English following~\citet{vec2text-google}, and (2) compressed text data (Compress) generated by asking an LLM to remove all redundant information from the sentences while preserving all semantic information. For both RTT and Compress, we use the original data as the input and the synthetic generation as the target. We provide more details in Appendix~\ref{appendix:dataset_generation}.

\textbf{Experimental Results.} We train 0.6B parameter multi-resolution AE models with the CRVQ quantizer at 9.63 bits per token using the same hyperparameters as in Section~\ref{sec:quant}. In addition to BLEU and the judge score, we report the ratio of numbers (NumKept) and proper nouns (PNounKept) that are matched between the original and reconstructed texts. To identify numbers and proper nouns during both training and evaluation, we use spaCy part-of-speech tags obtained using the \texttt{en\_core\_web\_sm} model~\citep{spacy}. In Table~\ref{table:supervision}, we observe that most of the proposed supervision methods do not beat the baseline using the standard CE loss with no masking on the Copy dataset. There are only two exceptions: encoder masking and the WeightedClass loss.
However the WeightedClass loss trades off better immutable token class performance for worse overall reconstruction. Encoder masking likely improves performance due to the fact that the model cannot solely rely on copying the tokens directly inside the latent representation, and instead learns better representations for predictions based on the context. We show additional encoder masking experiments in Table~\ref{table:encoder_mask_ratio} (Appendix~\ref{appendix:additional_experiments}). Lastly, we observe that, while RTT and Compress with the upscaler loss perform worse than the Copy dataset, decoder masking improves reconstruction, indicating that the upscaler loss is too rigid when the input and the target tokens do not match.

\begin{table}[htb!]
\caption{Supervision evaluation comparison of 0.6B parameter CRVQ multi-resolution AE models trained at 9.63 bits per token on 128 sequence length, scaling factor of 8, and latent dimension 24.}
\label{table:supervision}
\begin{center}
\resizebox{0.999\linewidth}{!}{
\small
\begin{tabular}{lrrrr}
\toprule
\multicolumn{1}{c}{\bf Supervision} &\multicolumn{1}{c}{\bf BLEU $\uparrow$}   &\multicolumn{1}{c}{\bf Judge Score $\uparrow$}    &\multicolumn{1}{c}{\bf NumKept $\uparrow$}    &\multicolumn{1}{c}{\bf PNounKept $\uparrow$} \\
\toprule
Copy Dataset, CE Loss, and No Mask & $\underline{0.76 \pm 0.01}$ & $\underline{3.25 \pm 0.14}$ & $0.91 \pm 0.01$ & $0.68 \pm 0.02$ \\
\toprule
\multicolumn{5}{l}{\textit{Loss Variants on Copy Dataset and No Mask}} \\
\midrule
BoundedCE$_{log2}$ & $0.45 \pm 0.01$ & $2.35 \pm 0.08$ & $0.51 \pm 0.02$ & $0.62 \pm 0.02$ \\
WeightedClass$_{10.0}$ & $0.61 \pm 0.01$ & $2.75 \pm 0.12$ & $\mathbf{0.94 \pm 0.01}$ & $\mathbf{0.83 \pm 0.01}$ \\
$\text{CE} + 0.1 \cdot \text{CE}_{LM}$ & $0.66 \pm 0.01$ & $2.62 \pm 0.12$ & $0.83 \pm 0.02$ & $0.60 \pm 0.02$ \\
\toprule
\multicolumn{5}{l}{\textit{Mask Variants on Copy Dataset and CE Loss}} \\
\midrule
Encoder Mask (5\%) & $\mathbf{0.78 \pm 0.01}$ & $\mathbf{3.49 \pm 0.15}$ & $\underline{0.93 \pm 0.01}$ & $\underline{0.71 \pm 0.01}$ \\
Decoder Mask (75\%) & $0.63 \pm 0.01$ & $2.89 \pm 0.15$ & $0.79 \pm 0.02$ & $0.58 \pm 0.02$ \\
Encoder (5\%) and Decoder (75\%) Mask & $0.62 \pm 0.01$ & $3.05 \pm 0.16$ & $0.77 \pm 0.02$ & $0.60 \pm 0.02$ \\
\toprule
\multicolumn{5}{l}{\textit{Dataset Variants on CE Loss}} \\
\midrule
RTT & $0.44 \pm 0.01$ & $2.35 \pm 0.12$ & $0.83 \pm 0.02$ & $0.43 \pm 0.02$ \\
\hspace{1em} with Decoder Mask (75\%) & $0.37 \pm 0.01$ & $2.83 \pm 0.12$ & $0.71 \pm 0.02$ & $0.44 \pm 0.02$  \\
Compress & $0.45 \pm 0.01$  & $2.42 \pm 0.13$  & $0.82 \pm 0.02$ & $0.51 \pm 0.02$ \\
\hspace{1em} with Decoder Mask (75\%) & $0.24 \pm 0.01$ & $2.71 \pm 0.11$ & $0.67 \pm 0.02$ &  $0.53 \pm 0.02$  \\
\bottomrule
\end{tabular}
}
\end{center}
\vspace{-1em}
\end{table}

\section{Evaluation}\label{sec:evaluation}

\subsection{Evaluation of Semantic Text Similarity}

\textbf{Limitations of Traditional Metrics.} We evaluate metrics commonly used in machine translation and text summarization for the purpose of evaluating semantic similarity between two texts. We compute BLEU~\citep{bleu}, chrF~\citep{chrf}, ROUGE-L~\citep{rouge}, and BertScore~\citep{bertscore} on a set of reference-hypothesis pairs as the mean of the per-sample metric scores. Additionally, we compute the ratio of numbers (NumKept) and proper nouns (PNounKept) matched between the reference and the hypothesis as in Section~\ref{sec:supervision}, as well as a Qwen3.6-35B-A3B-based~\citep{qwen36_35b_a3b} LLM judge score. We report the metrics across three datasets introduced in Section~\ref{sec:supervision} (Copy, RTT, and Compress), and the Corrupt dataset, where we replace all words assigned the verb part-of-speech tag in the hypothesis of the Copy dataset with the string "run", thereby corrupting information. Analogous to detection of numbers and proper nouns, we detect verbs using the same spaCy setup. Table~\ref{table:traditional_metrics} demonstrates that, while these metrics track linguistic reconstruction, they can give high scores even when the underlying information is destroyed, as seen in the Corrupt dataset. On the other hand, using the Qwen3.6-35B-A3B judge correctly assigns high scores to RTT and Compress datasets, while giving a low score to the Corrupt dataset, indicating that high scores on surface-level text similarity metrics do not imply semantic similarity.

\begin{table}[htb!]
\caption{Comparison of evaluation metrics on four datasets: Copy (identical reference and hypothesis), RTT (round-trip translation), Compress (redundancy removal), and Corrupt (verb replacement). We use the Qwen3.6-35B-A3B model as the judge.}
\label{table:traditional_metrics}
\begin{center}
\resizebox{\linewidth}{!}{
\small
\begin{tabular}{lrrrrrrr}
\toprule
\multicolumn{1}{c}{\bf Dataset}      &\multicolumn{1}{c}{\bf BLEU $\uparrow$}   &\multicolumn{1}{c}{\bf chrF $\uparrow$}     &\multicolumn{1}{c}{\bf Rouge-L $\uparrow$}  &\multicolumn{1}{c}{\bf NumKept $\uparrow$}  &\multicolumn{1}{c}{\bf PNounKept $\uparrow$}     &\multicolumn{1}{c}{\bf BERTScore $\uparrow$}   &\multicolumn{1}{c}{\bf Judge Score $\uparrow$}
\\
\midrule
Copy & $1.00 \pm 0.00$ & $1.00 \pm 0.00$ & $1.00 \pm 0.00$ & $1.00 \pm 0.00$ & $1.00 \pm 0.00$ & $1.00 \pm 0.00$ & $10.00 \pm 0.00$ \\
\midrule
RTT & $\underline{0.58 \pm 0.01}$ & $\underline{0.79 \pm 0.00}$ & $\underline{0.81 \pm 0.01}$ & $\underline{0.89 \pm 0.01}$ & $0.76 \pm 0.01$ & $\mathbf{0.79 \pm 0.01}$ & $\underline{9.42 \pm 0.06}$ \\
Compress & $0.44 \pm 0.02$ & $0.66 \pm 0.01$ & $0.70 \pm 0.01$ & $0.82 \pm 0.02$ & $\underline{0.90 \pm 0.01}$ & $0.60 \pm 0.02$ & $\mathbf{9.74 \pm 0.04}$ \\
Corrupt & $\mathbf{0.72 \pm 0.00}$ & $\mathbf{0.81 \pm 0.00}$ & $\mathbf{0.87 \pm 0.00}$ & $\mathbf{0.99 \pm 0.00}$ & $\mathbf{0.99 \pm 0.00}$ & $\underline{0.72 \pm 0.01}$ & $2.03 \pm 0.13$ \\
\bottomrule
\end{tabular}
}
\end{center}
\end{table}

\textbf{LLM Judge Formulation.} For the LLM-based judge, we adapt the Multidimensional Quality Metrics (MQM;~\citealt{mqm,mqm2}) used in machine translation to detect information mismatches, specifically semantic mismatches, over-specification, under-specification, unwarranted addition, and omission. From these rules, we ask the judge to determine the number of semantic information pieces in the reference and hypothesis texts, count the number of differences between the two, and assign a text similarity score between 1 (lowest) and 10 (highest). By using these criteria, the judge  assesses whether the reconstruction maintains the core meaning of the original text rather than just measuring surface-level overlaps. More details, including the prompt, are in Appendix~\ref{appendix:lm_judge_prompt}.

\begin{figure}[htb!]
\centering
\includegraphics[width=0.8\linewidth]{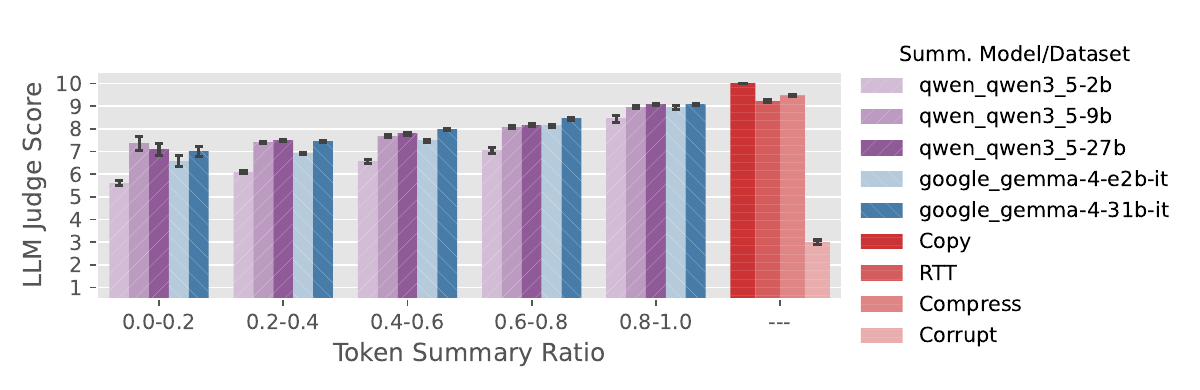}
\caption{LLM judge scores from the Qwen3.5-27B judge of text summaries using different summary models and target summary ratios, and four datasets: Copy (identical reference and hypothesis), RTT (round-trip translation), Compress (redundancy removal), and Corrupt (verb replacement).}
\label{fig:lm_judge_qwen}
\end{figure}

\textbf{LLM Judge Evaluation.} We test the monotonicity of the judge scores with respect to information overlap between the references and the hypotheses. We generate 30\%, 50\%, 70\%, and 90\% summary datasets using models from Qwen3.5~\citep{qwen35} and Gemma4~\citep{gemma4} model families of different sizes, and compute the token compression ratio for each of the summarized samples, using it as a proxy for ratio of information that was retained when summarizing. We compute the token compression ratio of each sample instead of using the target summary ratio because we find that the models do not always follow the target summary ratio. We show the results for the Qwen3.5-27B judge with thinking enabled on 1000 documents in Figure~\ref{fig:lm_judge_qwen}, highlighting that the judge scores monotonically increase both with the decrease in token compression ratio, as well as within the model family for larger models. To show that the judge generalizes across judge model families, we provide the results for the Gemma4-31B-it judge in Figure~\ref{fig:lm_judge_gemma} (Appendix~\ref{appendix:additional_experiments}).

\subsection{Text Compression Evaluation}\label{sec:compression}

\textbf{Compression and Reconstruction Comparison.} We compare our approach against lossless and lossy text compression methods. Namely, we compute compression and reconstruction performance of zlib~\citep{zlib}, bzip2~\citep{bzip2}, and zstandard~\citep{zstandard}, as provided by the Python standard library, the next-token probability distribution bitrate estimate of the Qwen3.5 model family~\citep{qwen35}, and the SONAR model estimate~\citep{sonar}. We train our multi-resolution AE models at 0.6B, 2.0B, and 4.3B scales using CRVQ with 2048 codebook size, 8 quantizers, scaling factor 8 (11 bits per token), latent dimension 48, 5\% encoder masking, and obtain the bitrate estimates using a 1.6B parameter autoregressive entropy model which we train on coarse-to-fine CRVQ code indices that the AE model outputs on the training data distribution. We obtain all bitrate estimates on a 22.82~MB sample of raw validation text data. Unless otherwise specified, for neural-based approaches, we use a sequence length of 128 tokens. In Table~\ref{table:baseline_and_scaling}, we see that our method performs close to zstandard in terms of compression rate, with both being outperformed by neural lossless compressors. Comparing neural lossy compression approaches, our models outperforms SONAR on both the compression rate and reconstruction performance, with larger models showing higher rate estimates, indicating higher-entropy code usage. While SONAR does not quantize its latent representations, making the bitrate inflated, even if such module was included and resulted in the same bitrate as our approach, our smallest model would still reconstruct text with higher BLEU and judge scores at 128 sequence length. Despite SONAR having a maximum sequence length of 512 tokens, actual usable length is shorter, as the performance drops significantly on longer sequences, likely due to its sentence-level training data.

\textbf{Improving Lossy Compression.} Although we achieve the best compression among neural lossy compression methods, their lossy nature should allow for bitrates lower than neural lossless methods. We hypothesize that the supervised learning objective is not enough to reach these rates and that reinforcement learning (RL) might be needed. However, running RL with multiple rollouts and LLM-based semantic similarity objective is expensive and outside the scope of this initial work. One RL variant could be to generate only two rollouts per input and using a trained binary scorer to gradually steer the model to better generations, which we leave as future work.

\begin{table}[htb!]
\caption{Comparison of compression rates (Bits per Byte) and reconstruction performance (BLEU and judge score) across traditional lossless, neural lossless, and neural lossy text compression methods. Unless otherwise specified, neural-based approaches use 128 sequence length.}
\label{table:baseline_and_scaling}
\begin{center}
\small
\begin{tabular}{lrrrr}
\toprule
\multicolumn{1}{c}{\bf Method}  &\multicolumn{1}{c}{\bf Bits per Token $\downarrow$}     &\multicolumn{1}{c}{\bf Bits per Byte $\downarrow$}   &\multicolumn{1}{c}{\bf BLEU $\uparrow$}     &\multicolumn{1}{c}{\bf Judge Score $\uparrow$}
\\
\toprule
\textit{Traditional Lossless} \\
\midrule
Zlib (Level 9) & $14.44$ & $3.05$ & $1.00 \pm 0.00$ & $10.00 \pm 0.00$ \\
Bzip2 (Level 9) & $11.36$ & $2.40$ & $1.00 \pm 0.00$ & $10.00 \pm 0.00$ \\
Zstandard (Level 3) & $13.91$ & $2.94$ & $1.00 \pm 0.00$ & $10.00 \pm 0.00$ \\
Zstandard (Level 22) & $10.50$ & $2.22$ & $1.00 \pm 0.00$ & $10.00 \pm 0.00$ \\
\toprule
\textit{Neural Lossless} \\
\midrule
Qwen3.5-0.8B-Base & $4.56$ & $0.96$ & $1.00 \pm 0.00$ & $10.00 \pm 0.00$ \\
Qwen3.5-2B-Base & $4.26$ & $0.90$ & $1.00 \pm 0.00$ & $10.00 \pm 0.00$ \\
Qwen3.5-4B-Base & $4.01$ & $0.85$ & $1.00 \pm 0.00$ & $10.00 \pm 0.00$ \\
\toprule
\textit{Neural Lossy} \\
\midrule
SONAR (32 seq. length) & $558.50$ & $120.10$ & $0.60 \pm 0.01$ & $8.33 \pm 0.10$ \\
SONAR (128 seq. length) & $140.76$ & $29.91$ & $0.15 \pm 0.01$ & $3.34 \pm 0.10$ \\
SONAR (512 seq. length) & $35.27$ & $7.47$ & $0.01 \pm 0.00$ & $1.56 \pm 0.04$ \\
Multi-resolution AE 0.6B & $9.93$ & $2.10$ & $0.87 \pm 0.01$ & $4.60 \pm 0.17$ \\
Multi-resolution AE 2.0B & $10.50$ & $2.22$ & $0.95 \pm 0.00$ & $7.27 \pm 0.17$ \\
Multi-resolution AE 4.3B & $10.62$ & $2.24$ & $0.98 \pm 0.00$ & $8.42 \pm 0.12$ \\
\bottomrule
\end{tabular}
\end{center}
\end{table}

\subsection{Downstream Task Evaluation}\label{sec:downstream_eval}

\textbf{Question-Answering Evaluation.} To evaluate downstream utility of the text reconstructions of our models, we formulate a downstream question-answering task based on MS MARCO dataset~\citep{msmarco}. We pass the combined text passages from the MS MARCO dataset~\citep{msmarco} through our text autoencoder models from Section~\ref{sec:compression} and SONAR at 128 sequence length, chunking and recombining sequences longer than 128 tokens. We then prompt the Qwen3.6-35B-A3B model to extract the answer from the reconstructed passage based on the MS MARCO question, and prompt the same model to compare the extracted answer to the true MS MARCO answer. Since the task relies on a model to extract the answer and to compare the answer, we compute an upper-bound score where the answer is extracted from an unmodified text passage (Same Passage Extraction). In Table~\ref{table:downstream_qa}, we observe that text reconstructions from our models reach the upper-bound score meaning that the text passage reconstructions retain information important in a question-answering task, while the SONAR model shows lower accuracy.

\textbf{Semantic Text Similarity Evaluation.} To evaluate the representation performance of our models, we compute the MTEB STSBenchmark score~\citep{mteb}. We evaluate two variants of our models, one with and one without the Qwen3-Embedding-0.6B~\citep{qwen3embedding} similarity matrix distillation, to the Qwen3-Embedding-0.6B and SONAR models. We use our text autoencoder architecture at 11 bits per token from Section~\ref{sec:compression} with one modification: since the outputs are chunked, we reshape them into a single vector and apply an MLP to obtain a single representation vector, which we also use for text reconstruction. For Qwen3-Embedding-0.6B similarity score distillation, we compute the similarity matrix on sentences sampled from the RTT dataset and use L2 loss to match the similarity scores. To measure reconstruction performance of Qwen3-Embedding-0.6B models, we train the upscaler and the decoder using the total scaling factor of 128 and the PRVQ quantizer with a codebook size 2048, 8 quantizers, and 16 subspaces (11 bits per token), with the rest of the model frozen. For all trained models, we use a hidden dimension of 1024 to match the hidden dimension of the Qwen3-Embedding-0.6B and SONAR models. In Figure~\ref{fig:downstream_emb}, we notice a representation-reconstruction trade-off, where our models perform well on the reconstruction task but do not provide representations useful for the semantic text similarity task, while SONAR and Qwen3-Embedding models perform well on the semantic text similarity task with lower text reconstruction performance. Additionally, even when training our models with similarity matrix distillation objective, while the STSBenchmark score increases from the random baseline, it remains low, which could result from the fact that autoencoder model representations generally perform worse compared to state-of-the-art representation learning approaches~\citep{tsdae}. Furthermore, we use a lower masking ratio than~\citet{tsdae}, which could further hinder representation learning.

\begin{table*}[htb!]
    \centering
    \begin{minipage}{0.41\textwidth}
        \caption{Accuracy of trained multi-resolution AE models at 11 bits per token, and SONAR on the MS MARCO-based question-answering task at 128 sequence length.}
        \label{table:downstream_qa}
        \vspace{-0.7em}
        \begin{center}
        \small
        \begin{tabular}{lr}
        \toprule
        \multicolumn{1}{c}{\bf Model}    &\multicolumn{1}{c}{\bf Accuracy $\uparrow$}
        \\
        \midrule
        Same Passage Extraction & $0.64 \pm 0.03$ \\
        \midrule
        SONAR & $0.55 \pm 0.03$ \\
        Multi-resolution AE 0.6B & $\underline{0.63 \pm 0.03}$ \\
        Multi-resolution AE 2.0B & $\mathbf{0.64 \pm 0.03}$ \\
        Multi-resolution AE 4.3B & $\mathbf{0.64 \pm 0.03}$ \\
        \bottomrule
        \end{tabular}
        \end{center}
    \end{minipage}%
    ~ \hspace{1em}
    \begin{minipage}{0.55\textwidth}
        \centering
        \includegraphics[width=\linewidth]{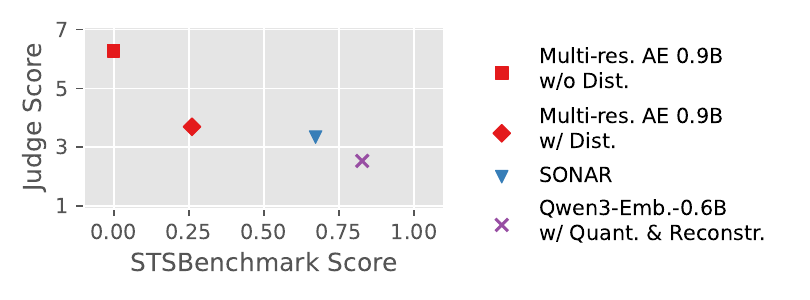}
        \captionof{figure}{Comparison of STSBenchmark (Spearman correlation) and BLEU score at 128 sequence length between multi-resolution AE at 11 bits per token, \mbox{Qwen3-Embedding} with text reconstruction at 11 bits per token, and full-precision SONAR.}
        \label{fig:downstream_emb}
    \end{minipage}%
\end{table*}

\section{Conclusion}
We have shown that text autoencoders can achieve low bitrates comparable to traditional lossless compression methods while reconstructing semantically equivalent text. Our findings indicate that these models prioritize copying information over learning high-quality representations, presenting an opportunity for future work to improve autoencoder representations, which could in turn improve the compression rate. Additionally, to match higher neural lossless compression rates, it might be necessary to train by rewarding semantically close but not identical reconstructions.
Given this, our proposed models are better suited for moderate compression and reconstruction over representation learning tasks, e.g., context compression or downstream tasks requiring reconstruction.

\ifarxiv
  \subsection*{Acknowledgments}

The authors are thankful to Skyler Seto for his help with datasets and Richard Bai for his suggestions and comments.
\fi

\bibliography{iclr2027_conference}
\ifarxiv
  \bibliographystyle{unsrtnat}
\fi

\newpage
\appendix
\crefalias{section}{appsec}
\crefalias{subsection}{appsec}

\section*{Appendix Contents}
\startcontents[appendix]
\printcontents[appendix]{}{1}{\setcounter{tocdepth}{2}}
\newpage
\section{LM Judge Scoring Examples}

We show scoring examples of references sampled from the validation set and hypotheses reconstructed from the references using our text autoencoder models, rated by the \mbox{Qwen3.6-35B-A3B}~\citep{qwen36_35b_a3b} judge.

\textbf{Judge Score:} 1\\
\textbf{Judge Reason:} Text 2 contains severe semantic mismatches, factual contradictions, and nonsensical substitutions across all its variants compared to Text 1. Key entities (Belgium/Russia), biological terms (eyed eggs/parr vs buffalo eggs/sei salmon/cow bits), statistics, and technical descriptions are entirely altered or fabricated. These changes completely distort the original scientific meaning and intent, rendering the texts semantically unrelated despite matching sentence structures. Therefore, the difference count is high and the global similarity score is minimal.\\
\textbf{Reference: }\\
Abstract\\
In the 1950s, Atlantic salmon had become extinct in the river Meuse basin, mainly owing to the construction of navigation weirs. In 1987 started a rehabilitation program of Atlantic salmon in Belgium. Stockings are being performed in six tributaries of the river Meuse, using eyed eggs, parr and pre-smolts obtained from foreign eggs reared in a hatchery of the 'Service de la Peche'. By the end of 1998, 896033 young salmon have been released in the wild, especially in the river Ourthe  basin. In most of the cases, the evaluation of stocking operations indicates good adaptation of parr in the tributaries where they are stocked, with autumnal 0+ parr densities up to 20-30 ind 100 m-2. However, fish mortality may occur when smolts pass through the turbines of the hydropower stations in the river Meuse (5\% per power plant), and most of the fish passes located downstream of Namur are not suitable for anadromous fish migration. The paper reports the results of stocking experiments, the evaluation of mortality rates in turbines and the  advancement of programmes aiming at the restoration of downstream and upstream migration (construction of fish migration facilities).\\
abstract = "In the 1950s, Atlantic salmon had become extinct in the river Meuse basin, mainly owing to the construction of navigation weirs. In 1987 started a rehabilitation program of Atlantic salmon in Belgium. Stockings are being performed in six tributaries of the river Meuse, using eyed eggs, parr and pre-smolts obtained from foreign eggs reared in a hatchery of the 'Service de la Peche'. By the end of 1998, 8 96033 young salmon have been released in the wild, especially in the river Ourthe basin. In most of the cases, the evaluation of stocking operations indicates good adaptation of parr in the tributaries where they are stocked, with autumnal 0+ parr densities up to 20-30 ind 100 m-2. However, fish mortality may occur when smolts pass through the turbines of the hydropower stations in the river Meuse (5\% per power plant), and most of the fish passes located downstream of Namur are not suitable for anadromous fish  migration. The paper reports the results of stocking experiments, the evaluation of mortality rates in turbines and the advancement of programmes aiming at the restoration of downstream and upstream migration (construction of fish migration facilities).",\\
N2 - In the 1950s, Atlantic salmon had become extinct in the river Meuse basin, mainly owing to the construction of navigation weirs. In 1987 started a rehabilitation program of Atlantic salmon in Belgium. Stockings are being performed in six tributaries of the river Meuse, using eyed eggs, parr and pre-smolts obtained from foreign eggs reared in a hatchery  of the 'Service de la Peche'. By the end of 1998, 896033 young salmon have been released in the wild, especially in the river Ourthe basin. In most of the cases, the evaluation of stocking operations indicates good adaptation of parr in the tributaries where they are stocked, with autumnal 0+ parr densities up to 20-30 ind 100 m-2. However, fish mortality may occur when smolts pass through the turbines of the hydropower stations in the river Meuse (5\% per power plant ), and most of the fish passes located downstream of Namur are not suitable for anadromous fish migration. The paper reports the results of stocking experiments, the evaluation of mortality rates in turbines and the advancement of programmes aiming at the restoration of downstream and upstream migration (construction of fish migration facilities).\\
AB - In the 1950s, Atlantic salmon had become extinct in the river Meuse basin, mainly owing to the construction of navigation weirs. In 1987 started a rehabilitation program of Atlantic salmon in Belgium. Stockings are being performed in six tributaries of the river Meuse, using eyed  eggs, parr and pre-smolts obtained from foreign eggs reared in a hatchery of the 'Service de la Peche'. By the end of 1998, 896033 young salmon have been released in the wild, especially in the river Ourthe basin. In most of the cases, the evaluation of stocking operations indicates good adaptation of parr in the tributaries where they are stocked, with autumnal 0+ parr densities up to 20-30 ind 100 m-2. However, fish mortality may occur when smolts pass through  the turbines of the hydrop\\
\textbf{Generation: }\\
Abstract\\
In the 1950s, Atlantic salmon had become rare in the river Meuse basin, mainly owing to the construction of navigation weirs. In 1987 started a rehabilitation program of Atlantic salmon in Russia. Stockings are being performed in six tributaries of the river Meuse to kill buffalo eggs, parr and pre-smet, obtained from foreign eggs reared in a hatchery of the 'Common de la Chiaco'. By the end of 1998, 896033 young salmon have been released in the wild, especially in the river Ourthe  basin. In each of the cases, the evaluation of stocking operations indicates good adaptation of parr in the tributaries where they are stocked, with autumnal 0+ parr densities up to 20-30 ind 100 m-2. However, fish mortality may occur when calabones pass through the islands of the hydropower stations in the river Meuse (i.e largest power plant), and most of the fish passes for downstream of Namur are not suitable for anadromous fish migration. The paper reports the results of stocking experiments, the evaluation of mortality rates in turbines and the  advancement of programmes aiming at the restoration of downstream and upstream migration (construction of fish migration facilities).\\
abstract = "In the 1950s, Atlantic salmon had become extinct in the river Meuse basin, mainly owing to the construction of navigation weirs. In 1987 started a rehabilitation program of Atlantic salmon in Belgium. Stockings are being performed in six tributaries of the river Meuse, using sei salmon, parr and test-stab, removed from annual eggs reared in a hatchery of the 'Service de la Peche'. By the end of 1998, 8 96033 young salmon have been released in the wild, especially in the river Ouria Alto. In most of the cases, the evaluation of stocking trends indicates good fertility of karr in the tributaries where they are stocked, with autumnal inrange zone karr densities up to 20-30 ind 100 n-2. However, fish mortality may occur when smolts pass through the turbines of the hydropower heater in the Mass Meuse (i.e major power plant), and most of the fish caught growing downstream of Namur are not suitable for anadromous fish  migration. The paper reports the results of stocking experiments, the decrease of mortality rates in trout and the advancement of programmes aiming at the restoration of downstream and upstream population (construction of fish migration policies).",\\
N2 - In the 1950s, Atlantic salmon had become extinct in the river Meuse basin, mainly owing to the construction of navigation weirs. In 1987 started a rehabilitation program of Atlantic salmon in Belgium. Stockings are being performed in six tributaries of the river Meuse, using cow bits, parr and pre-steins, collected from collecting chick Es in a hatchery  is the 'project in annual Peche'. By the end of 1998, 896033 young salmon have been released to the wild, especially in the river Ourum basin. In most of the cases, a reduction of stocking operations indicates good viability of parr in the aquashores where they are stocked, with fish per day and although parr densities up to 20-30 ind 100 m-2. However, fish mortality may occur when smolts pass through the ponds of the hydropower stations in the river JIP-5, main water body ), and most of the fish passes over the Bovur are not suitable for anadromous fish migration. The paper reports the results of stocking experiments, the evaluation of mortality rates in vitro and the advancement of programmes aiming at the restoration of lakes and tidal migration (construction of fish migration facilities).\\
AB - In the 1950s, Atlantic salmon had become extinct in the river Meuse salmon, mainly owing to the construction of navigation weirs. In 1987 started a rehabilitation program of Atlantic salmon in Belgium. Stockings are being performed in six tributaries of the river Meuse, using penc  eggs, parr and pre-smolts obtained from mature eggs reared in a hatchery of the 'Service of Annual Peche'. By the end of 1998, 806,33 young salmon have been released in the wild, especially in the river Ourous Woods. In most of the cases, the evaluation of stocking operations involved short waves of parr in the tributaries where they are stocked, with tubal 4 grams parr densities up to 20-30 ind 100 m-2. However, fish mortality had risen when smolts weighed through  the turbines of the hydrop\\
\\
\textbf{Judge Score:} 3\\
\textbf{Judge Reason:} There are 16 semantic differences, primarily factual distortions, name alterations, and meaning inversions. Key discrepancies include the treaty date changing from 303 BC to 1303 BC, "Indika" becoming "Polika", "Arrian" becoming "ArabH", multiple scholar names being altered, geographic/historical details shifting from rivers to wars, and critical quotes or interpretations being distorted or inverted (e.g., "finds faults only with" becomes "finds agreements only with"). These numerous factual errors and semantic mismatches severely compromise the historical accuracy and core meaning of the original text. Due to the extensive nature of these deviations, the global similarity score is lowered to 3.\\
\textbf{Reference: }\\
Megasthenes (350 BC - c. 290 BC) was an ancient Greek historian, diplomat and Indian ethnographer and explorer in the Hellenistic period. He described India in his book Indika, which is now lost, but has been partially reconstructed from the writings of the later authors. While Megasthenes's account of India has survived in the later works, little is known about him as a person, except that he was a Greek man. He must have been a learned man, as evident by the book he wrote. His appointment as an ambassador to India suggests that he must have been  a reputed officer by this time.\\
At the time of treaty between the Greek ruler Seleucus I Nicator and the Indian ruler Chandragupta Maurya in c. 303 BC, he appears to have been serving as an officer under Sibyrtius, who was Seleucus's satrap of Arachosia. According to Arrian, Megasthenes lived in Arachosia and travelled to Pataliputra. Megasthenes also visited the Maurya capital Pataliputra, but it is not certain which other parts of India he visited. He appears to have passed  through the Punjab region in north-western India, as he provides a detailed account of the rivers in this area. He must have then traveled to Pataliputra along the Yamuna and the Ganga rivers. Megasthenes compiled information about India in form of Indika, which is now a lost work, but survives in form of quotations by the later writers.\\
Among the ancient writers, Arrian (2nd century AD) is the only one who speaks favorably of Megasthenes. Diodorus (1st century BC) quotes Megasthenes by omitting some parts of his narratives. Other writers  explicitly criticize Megasthenes: Eratosthenes (2nd century BCE) accuses Megasthenes of engaging in falsehood, although he apparently borrowed much of his content about India from Megasthenes. Strabo (1st century AD) calls Megasthenes a liar for writing fabulous stories about India; he also brands as liars the other earlier writers on India, including Deimachus, Onesicritus, Nearchus. According to Strabo, "no faith whatever can be placed in Deimachos and Megasthenes". Pliny (1st century AD) criticizes  Megasthenes's description of the races of India, and his account of Herakles and Dionysus.\\
Modern scholars such as E. A. Schwanbeck, B. C. J. Timmer, and Truesdell Sparhawk Brown, have characterized Megasthenes as a generally reliable source of Indian history. Schwanbeck finds faults only with Megasthenes's description of the gods worshipped in India. Brown is more critical of Megasthenes, but notes that Megasthenes visited only a small part of India, and must have relied on others for his observations: some of  these observations seem to be erroneous, but others cannot be ignored by modern researchers. Thus, although he was often misled by the erroneous information provided by others, much of the information provided by him appears to be accurate, as evident by the fact that his work remained the principal source of information about India to the subsequent writers.\\
References:\\
Thomas C. Mcevilley (2012). The Shape of Ancient Thought: Comparative Studies in Greek and Indian Philosophies. Allworth. p. 538. ISBN 978-1-58115-933-2. Three Greek ambassadors  are known by name: Megasthenes, ambassador to Chandragupta; Deimachus, ambassador to Chandragupta's son Bindusara; and Dyonisius, whom Ptolemy Philadelphus sent to the court of Ashoka, Bindusara's son\\
\textbf{Generation: }\\
Megasthenes (350 BC - c. 290 BC) was an ancient Greek historian, diplomat, Indian ethnographer and explorer in the Hellenistic period. He described India in his book Indika, which would later become, but has been partially severed from the writings of the later authors. Although Megasthenes's account of India has survived in the later works, little is known about him as a person, except that he was a Greek man. He must have been a learned man, as evident by the book he wrote. His appointment as an ambassador to India suggests that he must have been  a reputed officer by this time.\\
At the time of treaty between the Greek ruler Seleucus I Nicator and the Indian ruler Chandragupta Maurya in c.1303 BC, he appears to have been serving as an officer under Sibyrtius, who was Seleucus's satrap of Arachosia. According to Arrian, Megasthenes lived in Arachosia and travelled to Pataliputra. Megasthenes also visited the Maurya capital Pataliputra, but it is not certain which other parts of India he visited. He appears to have passed  through the Punjab region in north-western India, as he provides a detailed account of the wars in this area. He describes thathe traveled to Pataliputra along the Yamuna and the Ganga rivers. Megasthenes compiled information about India in form of Polika, which is later a lost work, but survives in form of quotations by the later writers.\\
Like the ancient writers, Arabh (2nd century AD) is the only one who speaks favorably of Megasthenes. Diodorus (1st century BC) quotes Megasthenes by omitting some parts of his narratives. Other writers  explicitly criticize Megasthenes". Eratosthenes (2nd century BCE) accuses Megasthenes of engaging in falsehood, although he apparently borrowed much of his content about India from Megasthenes. Strabo (1st century AD) calls Megasthenes a liar for writing fabulous stories about India; he also treats as guilds the other earlier writers on India, including Deimachus,Engicritus, Nearchus. According to Strabo, "no religious works can be placed in Deimachus and Megasthenes". Pliny (1st century AD) criticizes  Megasthenes's description of the races of India, and his account of the hermits and Dionysus.\\
Modern scholars such as E. A. Schwanbeck, B. C. J. Rowmer, and Tecklins Sparker Brown, have characterized Megasthenes as a generally reliable source of Indian history. Schwanbeck finds agreements only with Megasthenes's description of the gods worshipped in India. Brown is more critical of Megasthenes, but notes that Megasthenes visited only a small part of India, and must have relied on others for his observations: some of  these observations seem to be erroneous, but others cannot be ignored by modern researchers. Thus, although he was often misled by the erroneous information provided by others, much of the information provided by him appears to be accurate, as evident by the fact that his work remained the principal source of information about India to the subsequent ones.
References:\\
Thomas C. Mina Cristian (2012). The Shape of Ancient Thought: Comparative Studies in Greek and Indian Philosophies. Allworth. p. 538. ISBN 978-1-58115-933-2. Three Greek ambassadors  are known by name: Megasthenes, ambassador to Chandragupta; Thimachus, ambassador to Chandragupta's son Bindusara; and Dyonisius, whom Ptolemy Philadelphus sent to the court of Ashoka, Bindusara's son\\
\\
\textbf{Judge Score:} 5\\
\textbf{Judge Reason:} Both texts contain approximately 9 core semantic information pieces related to profiling and performance tuning strategies. There are 8 distinct semantic differences, mostly caused by direct lexical substitutions that either alter technical meaning (e.g., 'instrumentation' to 'cloning', 'big' to 'right', 'eliminate' to 'reach') or introduce nonsensical/phrasal breaks ('they refuse' to 'whatever', 'dispartities' to 'sizeequalities', 'move on' to 'what was a day'). While the overarching structure and intent remain aligned, the high frequency of these substantive and readability-altering changes significantly degrades semantic equivalence, justifying a moderate global similarity score.\\
\textbf{Reference:}\\
since you can't create files, that rules out the hierarchical profiler, but the dbms\_profiler should work for you.\\
If you don't have the package (and tables) already installed, and you don't want to talk to your dba's or they refuse to install it, then you'll have to go with inline instrumentation of your own devising.\\
Do you have something running 10000000 times but each time running quickly?\\
So total time for that step is big? if so, you probably can't tune the individual run times but can you eliminate some executions?\\
conversely do you  ahve something running a few times but each time running long? If so, this is a good candidate for tuning.\\
last, do you have something with large total run time and with large dispartities between min/max/avg? If so, look to see why the code fluctuates so much. is there something you can do to eliminate the max end, or tune the max end so it runs more like the min?\\
if something runs 1000000 times but only consumes 10 seconds out of 25 minutes then who cares? move on to something else that is consuming a lot of time .\\
"A LOT" is relatvive. If the most time-consuming piece of your code only consumes 30 seconds, then that's your worse case and you need to look at that.\\
And you'll likely need to look at a lot of others since the execution is spread around\\
if the most time-consuming piece of your code consumes 24 minutes out of 25 then you've found your problem area, eliminate/reduce/tune that and you're likely done.\\
\textbf{Generation:}\\
since you can't create files, that rules out the hierarchical profiler, but the dbms\_profiler should work for you.\\
If you don't have the package (and tables) already installed, and you don't want to talk to your dba's or whatever refuse to install it, then you'll have to go with inline cloning of your own compressing.\\
Do you have something running 10000000 times but each time running quickly?\\
So total time for this step is right? if so, you probably can't tune the individual run times but can you reach some executions?\\
Conversely do you  ah, something running a few times but each time running long? If so, this is a good candidate for tuning.\\
last, do you have something with large total run time and with large sizeequalities between min/max/avg? If so, look to see why the code fluctuates so much. is there something you can do to eliminate the max end, or tune the max end so it runs more like the min?\\
if something runs 1000000 times but only consumes 10 seconds out of 25 minutes then who cares? what was a day else that is consuming a lot of time .\\
"A LOT" is relatvive. If the most time-consuming piece of your code only consumes 30 seconds, then that's your worse case and you need to look at that.\\
And you'll likely need to look at a lot of others since the execution is normal around\\
if the most time-consuming piece of your code consumes 24 minutes out of 25 then you've found your problem area, eliminate/reduce/tune that and you're likely done.\\
\\
\textbf{Judge Score:} 7\\
\textbf{Judge Reason:} Text 2 contains four semantic differences compared to Text 1: the replacement of "Critical Care" with "Patient Care" in the title and opening sentence, changing "bringing together" to "promoted through" in the third sentence, and substituting "Investigations" with "Connections" in the fourth sentence. These four changes modify the specific domain name and alter the described nature of the collaborative theme and its outputs. Despite these substitutions, over 60\% of the semantic content remains verbatim, indicating strong overall similarity but preventing a perfect score.\\
\textbf{Reference:}\\
Critical Care\\
Critical care involves the care of the sickest patients in the hospital. Critically ill patients have usually been through a significant insult to their body (such as trauma, infection, burn) and have developed organ failure and require life-support. Critical Care is the largest theme bringing together clinicians and scientists from diverse backgrounds and includes collaborative research from hospitals throughout north-west London. Investigations range from evaluating biological mechanisms of organ failure through to the development of innovative technologies which allow the short-term and long-term support and recovery of organs.\\
Many people are exposed to the environment of an Intensive care unit (ICU) either personally or through a  family member. It is often a life-changing event and our work aims to reduce this impact facilitating post-ICU recovery.\\
This data is extracted from the Web of Science and reproduced under a licence from Thomson Reuters. You may not copy or re-distribute this data in whole or in part without the written consent of the Science business of Thomson Reuters.\\
\textbf{Generation:}\\
Patient Care\\
Patient care involves the care of the sickest patients in the hospital. Critically ill patients have usually been through a significant insult to their body (such as trauma, infection, burn) and have developed organ failure and require life-support. Critical Care is the largest theme promoted through clinicians and scientists from diverse backgrounds and includes collaborative research from hospitals throughout north-west London.\\
Connections range from evaluating biological mechanisms of organ failure through to the development of innovative technologies which allow the short-term and long-term support and recovery of organs.\\
Many people are exposed to the environment of an Intensive care unit (ICU) either personally or through a  family member. It is often a life-changing event and our work aims to reduce this impact facilitating post-ICU recovery.\\
This data is extracted from the Web of Science and reproduced under a licence from Thomson Reuters. You may not copy or re-distribute this data in whole or in part without the written consent of the Science business of Thomson Reuters.\\
\\
\textbf{Judge Score:} 9\\
\textbf{Judge Reason:} The texts are nearly identical, sharing the same title, gameplay description, control instructions, URL, and closing recommendations. There is exactly one semantic difference: Text 1 states players will experience "new things and surprises," while Text 2 changes this to "hugging and surprises." This is a single lexical substitution that slightly alters a minor descriptive phrase but does not impact the core functional or narrative meaning. Other minor variations (e.g., "to" vs "and", omission of auxiliary verbs) are grammatical/stylistic and qualify as acceptable variation. Therefore, the difference count is 1 and the global similarity score is 9.\\
\textbf{Reference:}\\
Canoniac Launcher 2\\
You will feel a lot of fun and excitement when you join the online game Canoniac Launcher 2. You need to practice a lot of fascinating flying skills in the air. What are you waiting for? Let's play now and experience a lot of new things and surprises. Hit the start button and play hard. You can overcome everything. Our robot friend Jimmy is back again and you have to start him to get him to fly as far as possible.\\
This time you are on a strange planet full of aliens. Of course, there will be some cool upgrades that will make it more interesting. Play how  to get maximum score. Use mouse to aim. Hold down the left mouse button to launch. Release it to put out the fire. Click to shoot. You will play strong to shoot down the aliens. You will do a great job on that task. Let's conquer all things better.\\
Enjoy the online game Canoniac Launcher 2 at .\texttt{<redacted url>} and feel a lot of new points. Alien bad guys are everywhere. You need to fly high to shoot enemies. Do not land quickly or you will stop the game. Fly high and go as far as you can. You need to perform more high  flying skills. Keep the robot soar higher and enjoy the many experiences in the sky. Take on all new challenges!\\
The game has many levels. How many levels will you pass and how many scores will you get? It all depends on your ingenuity. Play now and share the fun with your friends. If you love it, invite your friends to join some other games similar to Stickman Huggy 2\\
\textbf{Generation:}\\
Canoniac Launcher 2\\
You will feel a lot of fun and excitement when you join the online game Canoniac Launcher 2. You need to practice a lot of fascinating flying skills in the air. What are you waiting for? Let's play now and experience a lot of hugging and surprises. Hit the start button and play hard. You can overcome everything. Our robot friend Jimmy is back again and you have to start him and get him to fly as far as possible.\\
This time you are on a strange planet full of aliens. Of course, there will be some cool upgrades that will make it more interesting. Play how  to get maximum score. Use mouse to aim. Hold down the left mouse button to launch. Release it to put out the fire. Click to shoot. You will play strong to shoot down the aliens. You will do a great job on that task. Let's conquer all things better.\\
Enjoy the online game Canoniac Launcher 2 at .\texttt{<redacted url>} and feel a lot of new points. Alien bad guys are everywhere. You need to fly high to shoot enemies. Do not land quickly or you will stop the game. Fly high and go as far as you can. You need to perform more high  flying skills. Keep the robot soar higher and enjoy the many experiences in the sky. Take on all new challenges!\\
The game has many levels. How many levels will you pass and how many scores will you get? It all depends on your ingenuity. Play now and share the fun with your friends. If you love it, invite your friends to join some other games similar to Stickman Huggy 2\\

\section{Additional Experiments}\label{appendix:additional_experiments}


\begin{figure}[htb!]
        \centering
        \includegraphics[width=0.33\linewidth]{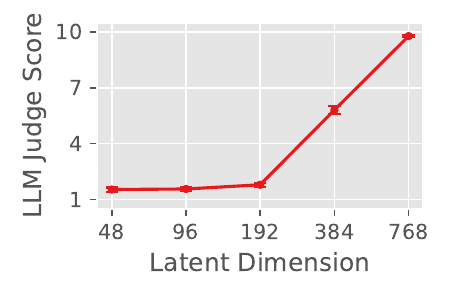}
        \caption{LLM judge score of 0.6B parameter multi-resolution AE models trained using the upscaler and the decoder loss at 128 sequence length, scaling factor of 128, hidden dimension 768, and varying latent bottleneck dimensions.}
        \label{fig:arch_cont_dim}
\end{figure}

\begin{figure}[htb!]
        \centering
        \includegraphics[width=0.33\linewidth]{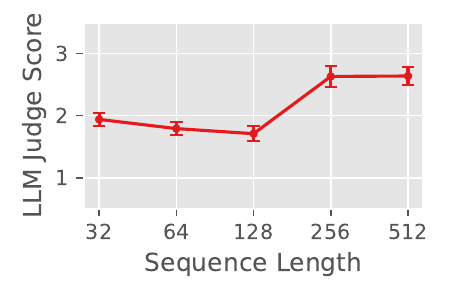}
        \caption{LLM judge score of 0.6B parameter multi-resolution AE models trained using the upscaler and the decoder loss, latent bottleneck dimension 48, scaling factor of 32, and varying sequence lengths.}
        \label{fig:arch_seq_len}
\end{figure}

\begin{table}[htb!]
\caption{BLEU and LLM judge score comparison of 0.6B parameter multi-resolution AE models trained on 128 sequence length, scaling factor of 128, and latent dimension 384 with varying upscaler and decoder loss weights.}
\label{table:upscaler_decoder_loss_comparison}
\begin{center}
\small
\begin{tabular}{lrr}
\toprule
\multicolumn{1}{c}{\bf Loss Weight}   &\multicolumn{1}{c}{\bf BLEU}   &\multicolumn{1}{c}{\bf Judge Score}
\\
\midrule
0.75 Upscaler + 0.25 Decoder & $\mathbf{0.92 \pm 0.01}$ & $\underline{5.68 \pm 0.18}$ \\
0.50 Upscaler + 0.50 Decoder & $\underline{0.91 \pm 0.01}$ & $\mathbf{5.78 \pm 0.20}$ \\
0.25 Upscaler + 0.75 Decoder & $0.66 \pm 0.02$ & $4.15 \pm 0.19$ \\
0.00 Upscaler + 1.00 Decoder &$ 0.77 \pm 0.01$ & $5.59 \pm 0.21$ \\
\bottomrule
\end{tabular}
\end{center}
\end{table}

\begin{table}[htb!]
\caption{BLEU and LLM judge score comparison of 0.6B parameter multi-resolution AE models trained on 128 sequence length with varying chunked residual vector quantization (CRVQ) configurations at 9.63 bits per token.}
\label{table:crvq_chunks}
\begin{center}
\resizebox{\linewidth}{!}{
\small
\begin{tabular}{rrrrrr}
\toprule
\multicolumn{1}{c}{\bf Codebook Size}  &\multicolumn{1}{c}{\bf Num. Quantizers}     &\multicolumn{1}{c}{\bf Scaling Factor} &\multicolumn{1}{c}{\bf Latent Dim.}    &\multicolumn{1}{c}{\bf BLEU}   &\multicolumn{1}{c}{\bf Judge Score}
\\
\midrule
2048 & 7 & 8 & 24 & $\mathbf{0.76 \pm 0.01}$ & $\mathbf{3.25 \pm 0.14}$ \\
2048 & 14 & 16 & 48 & $\underline{0.60 \pm 0.01}$ & $\underline{2.61 \pm 0.12}$ \\
2048 & 28 & 32 & 96 & $0.13 \pm 0.01$ & $1.56 \pm 0.11$ \\
2048 & 56 & 64 & 192 & $0.11 \pm 0.01$ & $1.56 \pm 0.11$ \\
\bottomrule
\end{tabular}
}
\end{center}
\end{table}

\begin{table}[htb!]
\caption{BLEU and LLM judge score comparison of 0.6B parameter CRVQ multi-resolution AE models trained on 128 sequence length, scaling factor of 8, and latent dimension 24, at 9.63 bits per token with varying encoder mask ratios at matched non-masked token count.}
\label{table:encoder_mask_ratio}
\begin{center}
\small
\begin{tabular}{rrr}
\toprule
\multicolumn{1}{c}{\bf Encoder Mask Ratio}   &\multicolumn{1}{c}{\bf BLEU}   &\multicolumn{1}{c}{\bf Judge Score}
\\
\midrule
0\% & $0.76 \pm 0.01$ & $3.25 \pm 0.14$ \\
5\% & $\underline{0.78 \pm 0.01}$ & $3.49 \pm 0.15$ \\
10\% & $\mathbf{0.79 \pm 0.01}$ & $\underline{3.68 \pm 0.16}$ \\
15\% & $\mathbf{0.79 \pm 0.01}$ & $\mathbf{3.72 \pm 0.16}$ \\
20\% & $\mathbf{0.79 \pm 0.01}$ & $\underline{3.68 \pm 0.16}$ \\
\bottomrule
\end{tabular}
\end{center}
\end{table}

\begin{figure}[ht!]
\centering
\includegraphics[width=0.8\linewidth]{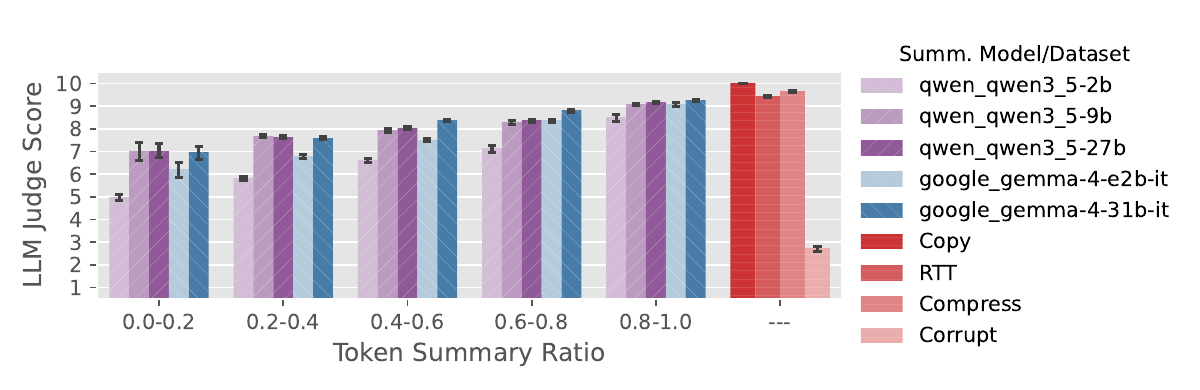}
\caption{LLM judge scores from the Gemma4-31B-it judge of text summaries using different summary models and target summary ratios, and four datasets: Copy (identical reference and hypothesis), RTT (round-trip translation), Compress (redundancy removal), and Corrupt (verb replacement).}
\label{fig:lm_judge_gemma}
\end{figure}

\section{Round-trip Translation and Compress Dataset Generation}\label{appendix:dataset_generation}

We generate the synthetic round-trip translation (RTT) and Compress datasets using the \mbox{Qwen3.6-35B-A3B} model with thinking disabled, and recommended hyperparameters for general purpose tasks with thinking disabled\footnote{\url{https://huggingface.co/Qwen/Qwen3.6-35B-A3B}}. RTT dataset generation process consists of several stages: (1) we ask the model to add sentence boundaries to the input text, (2) we ask the model to translate the data to German while preserving sentence boundaries, and (3) we ask the model to translate the German translation back to English while preserving sentence boundaries. Similarly, Compress dataset generation process consists of the following stages: (1) we ask the model to add sentence boundaries to the input text, and (2) we ask the model to remove redundant information while preserving semantic information and sentence boundaries. We use the sentence boundaries during training of our autoencoder models to match the input and target tokens by the longest matching sequence of original-synthetic pairs of sentences that can fit inside the context. For the dataset generation process, we use the same data that we use for training the models. We show the sentence boundary insertion, machine translation, and redundancy removal prompts.

\begin{graytitlebox}{Sentence Boundary Insertion Prompt}
\# TASK\\
Insert deterministic end-of-sentence markers into the provided text while preserving every other character exactly.\\
\\
\# RULES\\
1. Identify every fully terminated sentence.\\
2. Right after the terminal punctuation mark of each complete sentence, append the token: \#eos\_N\#, where N is a zero-based sequential counter of the current sentence (0, 1, 2, ...).\\
3. If the text ends mid-segment without a terminal punctuation mark, DO NOT append any \#eos\_N\# marker to it.\\
4. Preserve all original characters, spacing, line breaks, quotes, ellipses, and formatting exactly as provided. Do not normalize, merge, or delete anything except inserting the requested tokens.\\
5. Output ONLY the transformed text. Do not add introductions, explanations, markdown code blocks, or trailing newlines.\\

\# INPUT TEXT\\
\texttt{[INPUT TEXT]}
\end{graytitlebox}

\begin{graytitlebox}{Machine Translation Prompt}
\# TASK\\
Translate the following text into \texttt{[LANGUAGE]} with absolute preservation of all formatting, whitespace, and structural markers.\\
\\
\# STRICT INSTRUCTIONS\\
1. MARKER HANDLING: Each sentence in the source is immediately followed by a token formatted as `\#eos\_N\#` (where N starts at 0 and increments sequentially). Retain every `\#eos\_N\#` token in its EXACT original position. Do NOT translate, remove, reorder, modify, recalculate, or omit them.\\
2. STRUCTURAL FIDELITY: Preserve all newlines, indentation (tabs/spaces), paragraph breaks, punctuation, dashes, brackets, quotes, and spacing character-for-character.\\
3. TRANSLATION SCOPE: Translate only the actual linguistic content. Non-linguistic elements and `\#eos\_N\#` markers must pass through completely unchanged.\\
4. OUTPUT CONTROL: Return ONLY the final translated text. Do not include greetings, explanations, markdown wrappers, or any extra commentary.\\
\\
\# INPUT TEXT\\
\texttt{[INPUT TEXT]}\\
\end{graytitlebox}

\begin{graytitlebox}{Redundancy Removal Prompt}
\# TASK\\
You are a precision compression engine.\\
Transform the following text into a maximally compressed version that preserves 100\% of the semantic content, narrative, tone, voice, and stylistic essence. You must not output exactly the same text, your output must be compressed.\\
\\
\# STRICT INSTRUCTIONS:\\
1. Keep ALL exact quotes, dates, names, places, technical terms, and numerical data unchanged.\\
2. Remove: redundant phrases, verbose transitions, filler words, rhetorical padding, repetitive clauses.\\
3. Paraphrase when it reduces length while preserving meaning.\\
4. Preserve tone/style (e.g., formal, ironic, poetic, technical, conversational, passive/active voice) through precise word choice and sentence rhythm.\\
5. Do not add, infer, or omit semantic information.\\
6. Each sentence in the input text is immediately followed by a token formatted as `\#eos\_N\#` (where N starts at 0 and increments sequentially). Retain every `\#eos\_N\#` token in its EXACT original position. Do NOT remove, reorder, modify, recalculate, or omit them. Keep the exact same number of sentences in the compressed version, with each sentence being compressed version of the original sentence. Sentence separators \#eos\_N\# between the input and the original must match.\\
7. Output ONLY the compressed text. No explanations, no markdown.\\

\# INPUT TEXT\\
\texttt{[INPUT TEXT]}\\
\end{graytitlebox}

\section{LM Judge Prompt for Evaluation of Semantic Matching}\label{appendix:lm_judge_prompt}

We show the LLM judge prompt used for the judge score evaluation. When running the judge, we enable thinking mode and use the recommended sampling hyperparameters for general purpose tasks with thinking mode enabled\footnote{\url{https://huggingface.co/Qwen/Qwen3.5-27B}}\footnote{\url{https://huggingface.co/Qwen/Qwen3.6-35B-A3B}}\footnote{\url{https://huggingface.co/google/gemma-4-31B-it}}.

\begin{graytitlebox}{LM Judge Prompt}
\# TASK

Determine the semantic difference of the two provided texts. Your main output is two different integer scores. First output is the number of differences where a score of 0 means that the texts do not contain any semantic differences and higher score means that the semantic difference is higher. Second is the global score from 1 to 10 where a lower score means that the texts are more different and a higher score means that the texts are more similar.
\\
\\
\# NUMBER OF DIFFERENCES RULES

1. Determine the semantic difference between the two texts. Semantic difference means the number of differences between the texts.

2. For each text, count the number of semantic information pieces that it contains. A semantic information piece might be a name, a fact, a year, or the fact that the text contains a greeting.

3. Compare the differences between the texts. For each difference (either text contains a semantic piece of information that the other doesn't, or the semantic information differs), add 1 to the total score, starting from 0.

4. You must not punish the fact that the text may be written differently, you must compare the semantic information present in the text.
\\
\\
\# GLOBAL SCORE RULES

1. The global score must be from 1 to 10 and correspond to how well the semantic information in the two texts matches.

2. Give a lower score if there are any differences which make the two texts significantly more different. Punish factual differences more than similar meaning expressed through different phrasing.

3. Give a higher score if the texts carry only small differences with similar meanings.

4. Give the minimum score of 1 if the texts are completely different or unrelated and give the maximum score of 10 if the two texts do not contain any semantic information differences.
\\
\\
\# SEMANTIC DIFFERENCE CATEGORIES \& GUIDELINES

Use the following categories to identify and classify semantic differences between Text 1 (original) and Text 2 (approximation). These categories will help you accurately count differences and calibrate the global similarity score. Do not output these category names in your JSON; use them internally to inform your judgment.

\#\# 1. Semantic Mismatch (Factual/Intent Distortion)

- **What it is:** Text 2 conveys a meaning that directly contradicts, inverts, or significantly distorts a core fact, negation, number, relationship, or intent from Text 1.

- **Bad Example:** Text 1: "The drug must NOT be taken with alcohol." Text 2: "The drug should be taken with alcohol."

- **Good Example:** Text 1: "The system requires a password." Text 2: "Authentication via a password is mandatory." (Preserves exact meaning)

\#\# 2. Over-specification

- **What it is:** Text 2 adds unnecessary constraints, narrows the scope, or specifies details not present in Text 1, potentially altering applicability.

- **Bad Example:** Text 1: "Users should update their browsers regularly." Text 2: "Chrome users must update their browsers every week."

- **Good Example:** Text 1: "Update your software." Text 2: "Ensure your software is up to date." (Equivalent scope)

\#\# 3. Under-specification

- **What it is:** Text 2 generalizes or removes specific details present in Text 1, making the output less precise or broader than intended.

- **Bad Example:** Text 1: "The committee approved the Q3 marketing budget." Text 2: "The committee approved a budget."

- **Good Example:** Text 1: "The Q3 marketing budget was approved." Text 2: "A budget was approved for Q3 marketing." (Equivalent precision)

\#\# 4. Unwarranted Addition (Hallucination)

- **What it is:** Text 2 introduces new claims, facts, names, dates, or relationships entirely absent from Text 1.

- **Bad Example:** Text 1: "The launch is scheduled for Friday." Text 2: "The launch is scheduled for Friday, and the CEO will attend."

- **Good Example:** Text 1: "The launch is Friday." Text 2: "Friday is the planned launch date." (No new info)

\#\# 5. Omission

- **What it is:** Text 2 fails to include semantic information that is clearly present in Text 1.

- **Bad Example:** Text 1: "The algorithm reduces latency by 30\% and improves accuracy by 15\%." Text 2: "The algorithm reduces latency."

- **Good Example:** Text 1: "It reduces latency and improves accuracy." Text 2: "Both latency reduction and accuracy improvement are achieved." (No loss)

\#\# Acceptable Variation (Do NOT count as differences)

- Synonymous phrasing, grammatical restructuring, stylistic changes, or reordering of clauses that preserve the exact semantic content.

- Minor formatting, punctuation, or capitalization differences.

- Contextually implied information that does not change the core meaning or scope.
\\
\\
\# OUTPUT RULES

1. Show your choice in plaintext json format containing the number of semantic information in both texts, the given scores and the reason in the reason field, e.g., \{"num\_info\_text\_1": 2, "num\_info\_text\_2": 4, "num\_differences": 1, "global\_score": 8, "reason": "The reason why the number of differences is 1 and why the global score is 8."\}. Do not output any other greetings, reasoning or markdown.

2. Ensure the JSON is valid and strictly follows the requested key names.
\\
\\
\# INPUT TEXT 1

\texttt{[INPUT TEXT]}
\\
\\
\# INPUT TEXT 2

\texttt{[RECONSTRUCTED TEXT]}
\end{graytitlebox}

\end{document}